\documentclass{article}

\usepackage{microtype}
\usepackage{graphicx}
\usepackage{subfigure}
\usepackage{booktabs} 

\usepackage{hyperref}

\newcommand{\tabincell}[2]{\begin{tabular}{@{}#1@{}}#2\end{tabular}}  

\usepackage[accepted]{icml2021}
\usepackage{booktabs}
\usepackage{multirow}
\usepackage{siunitx}
\usepackage{amsfonts}
\usepackage{xspace}

\icmltitlerunning{Graph Contrastive Pre-training for Effective Theorem Reasoning}

\begin{document}

\twocolumn[
\icmltitle{Graph Contrastive Pre-training for Effective Theorem Reasoning}



\icmlsetsymbol{equal}{*}

\begin{icmlauthorlist}
\icmlauthor{Zhaoyu Li}{McGill,Mila}
\icmlauthor{Binghong Chen}{Gatech}
\icmlauthor{Xujie Si}{McGill,Mila,cifar}
\end{icmlauthorlist}

\icmlaffiliation{Gatech}{Georgia Institute of Technology}
\icmlaffiliation{McGill}{McGill University}
\icmlaffiliation{Mila}{Mila - Quebec AI Institute}
\icmlaffiliation{cifar}{CIFAR AI Research Chair}
\icmlcorrespondingauthor{Xujie Si}{xsi@cs.mcgill.ca}

\icmlkeywords{Machine Learning, Reasoning, Theorem Proving, Graph Neural Network, Contrastive Learning}

\vskip 0.3in
]



\printAffiliationsAndNotice{}  

\newcommand{\tool}{NeuroTactic\xspace}

\begin{abstract}
Interactive theorem proving is a challenging and tedious process, which requires non-trivial expertise and detailed low-level instructions (or tactics) from human experts. Tactic prediction is a natural way to automate this process.
Existing methods show promising results on tactic prediction by learning a deep neural network (DNN) based model from proofs written by human experts. 
In this paper, we propose \tool, a novel extension with a special focus on improving the representation learning for theorem proving. \tool leverages graph neural networks (GNNs) to represent the theorems and premises, and applies graph contrastive learning for pre-training. We demonstrate that the representation learning of theorems is essential to predict tactics. Compared with other methods, \tool achieves state-of-the-art performance on the CoqGym dataset.
%

\end{abstract}

\section{Introduction}

\begin{figure}[ht]
\begin{center}
\centerline{\includegraphics[width=\columnwidth]{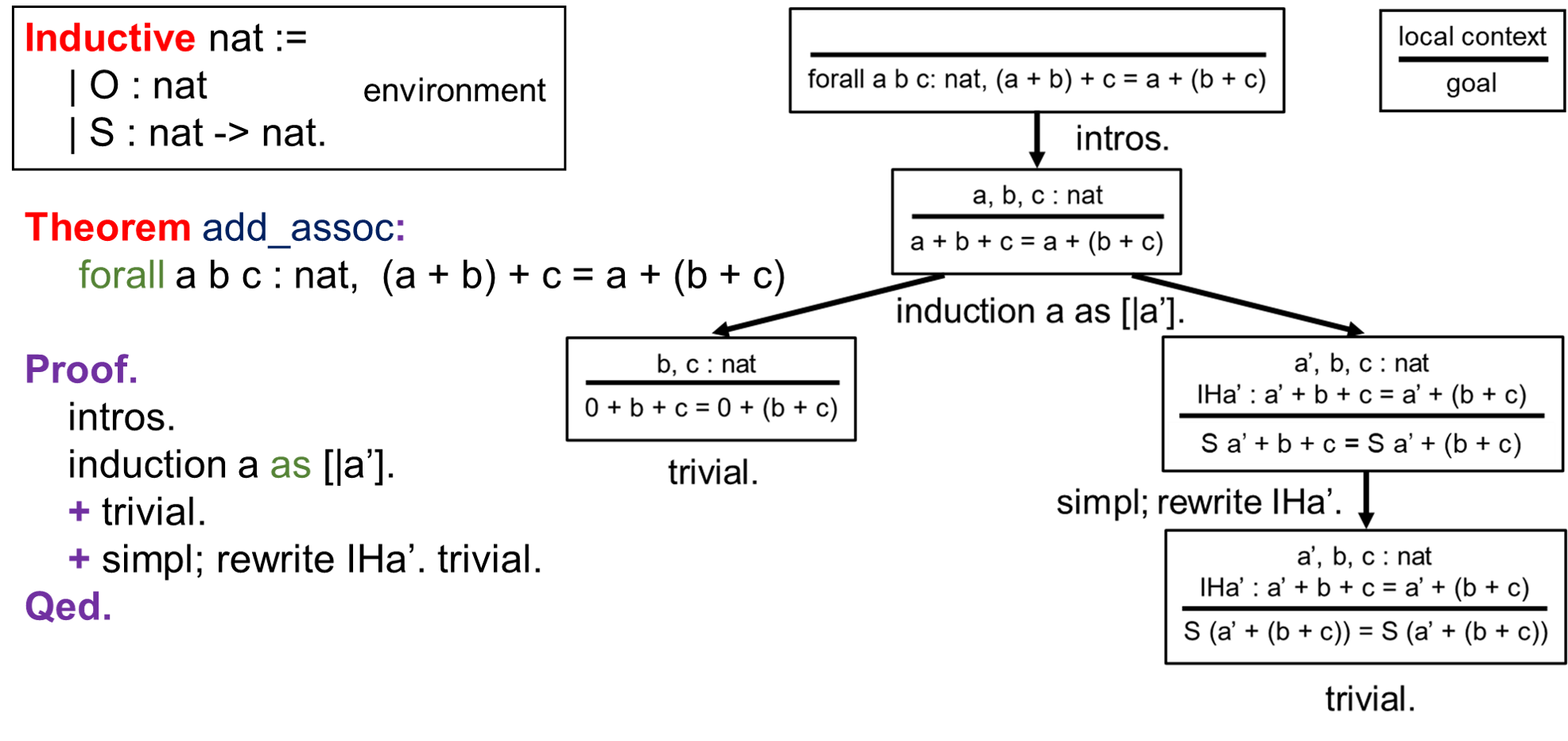}}
\vskip -0.1in
\caption{An example of using tactics to prove associativity of addition in Coq proof assistant. Given the mathematical definitions in the environment and the theorem to prove, a human expert provides a sequence of tactics (\textit{left}), performing high-order reasoning based on the semantic and structure information of this theorem. This process in Coq will generate a proof tree (\textit{right}), where each node in this tree shares the same environment, has a goal to decompose and a unique local context with practical premises to use.}
\label{figure1}
\end{center}
\vskip -0.4in
\end{figure}

Automated reasoning over mathematics proofs is an intriguing challenge for artificial intelligence, as it often requires machines to understand sophisticated high-order logic for reasoning. Interactive theorem proving (ITP) ~\citep{harrison2014history} is created to tackle this issue, which allows humans to develop formal proofs of mathematical theorems by interacting with a computer system. In ITP, human experts can define mathematical objects, state theorems, and prove them by entering a sequence of commands called tactics. As shown in Figure~\ref{figure1}, tactics decompose the current goal into several sub-goals that remained to be proved (0 sub-goals means that the current goal has been proved).  A successful proof script is a series of tactics that decompose the theorem completely with no sub-goals left. In this paper, we study a popular ITP, Coq proof assistant \cite{barras1997coq}
.

Although ITPs like Coq have achieved remarkable successes in software verification, the development process is usually labor-intensive and requires non-trivial expertise. For example, the CompCert compiler certification project \cite{leroy2016compcert} takes six PhD-years, which includes more than 100,000 lines of proof script. Nevertheless, these verification projects have accumulated a significant amount of valuable data carefully produced by human experts, which provides a promising opportunity for learning-based ITP. A number of recent works \cite{yang2019learning, sanchez2020generating, first2020tactok} apply machine learning methods to automate Coq proofs. These attempts focus on directly predicting a sequence of appropriate tactics that presumably prove the given theorem. However, a human expert rarely writes a sequence of tactics directly, although that is the final outcome. Instead, a human expert usually speculates a high-level plan at first before writing down any tactics. An important part of such a high-level plan is to figure out lemmas or premises that are going to be used. In this work, we leverage this insight for better representation learning of theorem reasoning in Coq.

Recently, pre-training for learning better representations has been used to improve the downstream tasks in many fields, like vision \cite{chen2020simple, he2020momentum} and graphs \cite{zhu2020deep, qiu2020gcc, thakoor2021bootstrapped}. These approaches perform contrastive learning in some pretext tasks, forcing representations of related objects to be close to each other and keeping apart of representations of unrelated ones. 
Inspired by these recent successes, we develop a contrastive-learning based pre-training strategy to learn the effective representations for theorem proving, which incorporates both semantic and structural information that is necessary for high-order logical reasoning.





In this work, we propose \tool, a novel contrastive learning-based approach to automating Coq proofs. Specifically, we design an appropriate pretext task, namely premise selection, and develop a graph contrastive learning approach.
We then exploit the learned representations for the downstream task, tactic prediction. We evaluate our approach with the benchmark dataset CoqGym. The experimental results show that \tool significantly outperforms the state-of-the-art model with relatively 16.2\% improvement.


\section{Related Work}

\textbf{Tactic prediction} Several recent works have developed machine learning methods for ITP. These methods employ supervised learning from existing proofs written by human experts in ITP. CoqGym \cite{yang2019learning} is most related to our work. CoqGym constructs a large-scale dataset and learning environment for Coq. It also develops ASTactic, a deep learning model that has an encoder-decoder architecture. The encoder embeds Coq terms in the form of ASTs by TreeLSTM \cite{tai2015improved}. The decoder generates program-structured tactics by expanding an AST based on its tactic grammar. TacTok \cite{first2020tactok} leverages previous proof scripts to improve tactic reasoning. ProverBot9001 \cite{sanchez2020generating} devises a simple model using MLPs and RNNs to predict the tactic and its arguments at each step and learns the proof in the CompCert project. \cite{paliwal2020graph} proposes the first use of GNNs to represent theorems in HOList ITP, which demonstrates the expressiveness of GNNs for theorem proving.

\textbf{Premise selection} Premise selection is to select relevant statements that are useful for proving a given theorem. There are also methods concerning this task in other ITPs. HOLStep \cite{kaliszyk2017holstep} provides a dataset for premise selection based on HOL Light ITP. It designs a CNN-LSTM model to encode tokens or characters in the theorems and premises. FormulaNet \cite{wang2017premise} converts statements in the HOLStep into a graph structure and applies a novel graph embedding that can preserve the order of edge to represent formulas.  However, these works focus on premise selection alone, while we use it as a pre-training task with contrastive learning to perform better high-order logic reasoning based on the learned representations.

\textbf{First-order theorem proving} Other works \cite{loos2017deep, kaliszyk2018reinforcement} focus on developing machine learning methods for theorem proving using first-order logic. However, these methods represent theorems and proofs in a fairly low-level language and are hard to accomplish high-level mathematical reasoning like proving theorems in Coq.


\section{Method}
\subsection{Framework Overview}
\begin{figure*}[ht]
\begin{center}
\centerline{\includegraphics[width=\columnwidth*2]{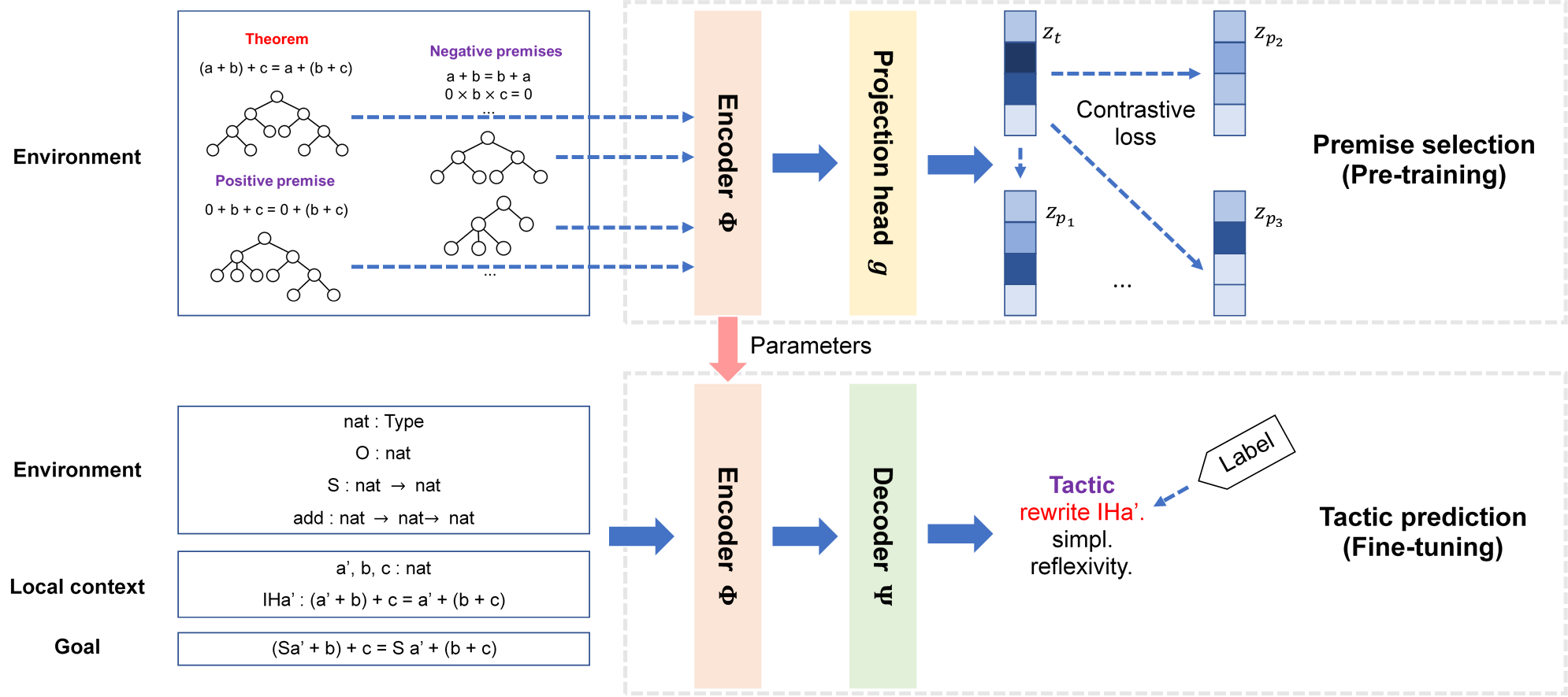}}
\caption{The overview of \tool's framework. The pretext task of the pre-training stage is premise selection, and the downstream task of the fine-tuning stage is tactic prediction.}
\label{figure2}
\end{center}
\vskip -0.4in
\end{figure*}

\tool consists of two stages, namely \textit{pre-training} stage and \textit{fine-tuning} stage, which are illustrated in Figure~\ref{figure2}. Although necessary, the supervision for the pre-training stage comes for free, since it is strictly a subset of the supervision used for the downstream task. More concretely, positive premises used in the pre-training stage are part of the parameters of tactics, which are the supervision in the fine-tuning stage. The intuition why such a pre-training stage is helpful is because premises are important hints for high-level planning and the pre-training stage encourages representation learning to pay particular ``attention'' to premises, which can be viewed as an indirect way for better planning.



\subsection{Semantic-guided Graph Contrastive Pre-training}

\textbf{Theorem representations} 
With dependent types and various syntax sugars, Coq allows human users to specify theorem definitions in a very flexible manner, which however makes it hard to have a uniform representation of all possible theorems at the surface level.
Following the method in CoqGym, we use SerAPI \cite{arias2016serapi} to extract the kernel-level representations of theorems, where we can have relatively simple grammar. 
Thus, theorems are represented by abstract syntax trees (ASTs) at the kernel level, rather than text as depicted in Figure~\ref{figure1}.

\textbf{Theorem encoding} 
Many previous works encode ASTs into vectors by TreeLSTM \cite{tai2015improved}. 
However, TreeLSTM fails to consider the information of the parent node and the sibling nodes when updating each node. Instead, we treat the ASTs as undirected graphs and use a GNN as our encoder $\Phi$ to map these graphs into a vector. Specifically, each AST can be viewed as a graph $G = (V, E)$ with a set of nodes $V$ and a set of edges $E$. The initial node attributes of node $v \in V$ is the one-hot encoding of its syntactical role in the kernel-level grammar, which is indicated by $x_v$. We employ the form in GIN \cite{xu2018powerful} to update each node $v$ in an AST in the following manner:
\begin{equation}
    h_v^{(k)} = {\rm MLP}\ (h_v^{(k-1)} + \sum_{u \in N(v)} h_{u}^{(k-1)}; \theta_k )
\end{equation}
where $N(v)$ is the neighbors of node $v$, $h_{v}^{(0)} = x_{v}$, and $h_{v}^{(k)}$ is the representation vector of node $v$ at the $k$-th iteration. Following GIN, the graph representation $h_G$ is the aggregation of each node in each iteration. We use mean pooling to aggregate all node representations for each iteration $1, \cdots, K$, and take the average of these $K$ vectors to obtain the entire graph representation $h_G$ of an AST.

\textbf{Semantic-guided training} Whether a premise in the environment or local context is relevant or not often depends on its semantic relevance with the given goal or theorem to prove. Figure~\ref{figure2} shows an interesting example -- the premise $\rm {IHa'}$ has the same semantic components ($\rm a'$, $\rm b$, $\rm c$ and addition) with the current goal and can serve as a tactic argument for proving it. Many practical premises in the local context share similar or even the same sub-expressions of the given goal. Applying these useful premises as tactic arguments helps decompose the current goal and simplify the proof. Based on these observations, we adopt a graph contrastive learning approach to pre-training effective representations. Previous works apply transformations on node features or graph structures, or sample sub-graphs to generate different views of graphs. However, these approaches are not applicable to our setting, since randomly generated AST samples are unlikely to obey the kernel-level grammar. To address this challenge, we leverage premise selection as our pretext task, employing the existing theorems and premises as our learning pairs and guiding the contrastive learning procedure with the semantic similarity.
Theorems and practical premises with similar semantics can serve as positive samples.The unrelated but similar structured premises can also be treated as hard negative samples. 

Specifically, given the current theorem $t$ and $N$ premises in the environment or local context, which are ASTs $a_t$, $a_{p_1}$, $\cdots$, $a_{p_N}$, we obtain their neural representations by applying an encoder $\Phi$ so that  
$h_t=\Phi(a_t)$, $h_{p_i}=\Phi(a_{p_i})$, $i=1,\cdots,N$. Without loss of generality, let's suppose $p_1$ is the positive sample while other premises are negative ones. We first apply a projection head $g(\cdot)$ on these representations to map them into a latent space: $z_t = g(h_t)$, $z_{p_i} = g(h_{p_i})$, $i=1,\cdots,N$. Then we leverage a contrastive loss for training. Specifically, we adopt InfoNCE loss \cite{oord2018representation}:
\begin{equation}
    \mathcal{L}\ (\Phi,\ g) = -\ {\rm log}\ \frac{{\rm exp}\ (z_t^{\top} z_{p_1})}{\sum_{i=1}^{N} {\rm exp}\ (z_t^{\top} z_{p_i})}
\end{equation}

InfoNCE maximizes the mutual information between positive pairs: it encourages the representation of the theorems and the positive premises being close to each other and being further away from negative samples in the latent space.

\textbf{Inference} To evaluate the pre-training performance of premise selection, we employ the same architecture during training stage and select the most relevant premise $p^{*}$ among the $N$ premises: $p^{*} = \arg\max_{p_i}(z_t^{\top}z_{p_i})$, $i=1,\cdots,N$.

\subsection{Fine-tuning for Tactic Prediction}

As illustrated in Figure~\ref{figure2}, \tool adopts an encoder-decoder architecture for tactic prediction and train the whole neural network with the ground-truth tactic label at each proof step. Given the premises and the goal expressed as ASTs, the encoder $\Phi$ embeds them into vectors. To leverage the learned representations in premise selection, we use the pre-trained parameters as the initialization of the encoder. Then we employ the same decoder $\Psi$ in ASTactic \cite{yang2019learning}: conditioned on embeddings from the encoder, the decoder generates tactic and its arguments by expanding an AST following a designed context-free grammar (CFG) for its tactic space. 

Specifically, let $p_t$ be the probabilities of all valid production rules in the CFG at $t$ expanding step after applying a $\textit{softmax}$, $y_t$ be the one-hot label of the ground-truth action, the framework is optimized by applying the cross-entropy loss with teacher forcing \cite{williams1989learning} to maximize the likelihood of ground-truth action at each step:
\begin{equation}
    \mathcal{L}\ (\Phi,\ \Psi) = -\mathbb{E}_t\ \left[ \sum_i y_{ti}\log\ (p_{ti}) \right]
\end{equation}

\section{Experiments}
\subsection{Datesets}
We evaluate our framework on the benchmark dataset CoqGym, which includes 71k human-written proofs from 123 Coq projects. There are 70 projects for training, 26 projects for validation, and 27 projects for testing. CoqGym extracts proof steps following its tactic grammar for tactic prediction. However, to our best knowledge, there is no dataset for premise selection using Coq proof assistant. To this end, based on CoqGym, we construct PremiseGym, the first dataset for premise selection in Coq. We extract the practical premises that have been used as tactic arguments in the proof as positive samples. For negative samples, since human experts always state relevant theorems in sequential order, we choose up to 10 the closest premises in the same Coq file before the given theorem as hard negative samples. Each instance in PremiseGym includes a theorem, a positive premise, and more than 8 negative premises on average.


\subsection{Settings}
For baseline model ASTactic, we set 256-dimensional vectors for all embeddings in encoder and decoder and use the checkpoint provided in \cite{yang2019learning}. For our encoder $\Phi$, 5 GNN layers are applied and the hidden units are of size 256. MLP is a two-layer fully-connected network with batch normalization and ReLU as activation function. The projection head is a 2-layer MLP in our implementation.

We train our model via the Adam optimizer with a learning rate of $1 \times 10^{-3}$. The epoch of premise selection is 20 and that of tactic prediction is 5.

\subsection{Experiment Results}



\textbf{Effect of contrastive pre-training} Table~\ref{table1} presents the performance of tactic prediction on the CoqGym dataset. Our approach outperforms ASTactic on most of the projects and improves the overall accuracy from 18.20\% (14,287/78,494) to \textbf{21.15\%} (\textbf{16602}/78494), more than \textbf{16.2\%} relative improvement. The performance of \tool demonstrates the effectiveness of our contrastive pre-training strategy and shows the huge impact the graph representations of Coq terms have on high-order reasoning for synthesizing proofs.

\begin{table}[t]
\vskip -0.1in
\caption{Correct number for tactic prediction on CoqGym testing set. ASTactic employs TreeLSTM as encoder, \tool adopts GIN as encoder and applies graph contrasitve learing (GCL) for pre-training. We also replace GIN to TreeLSTM for ablation study.}
\vskip -0.1in
\label{table1}
\begin{center}
\begin{small}
\begin{tabular}{l|r|r|r|r}
\toprule
\multirow{2}{*}{Project} & ASTactic & \multicolumn{2}{c|}{\tool} & \multirow{2}{*}{Total}\\
& {\tiny TreeLSTM} & {\tiny TreeLSTM+GCL} & {\tiny GIN+GCL} & \\
\midrule
PolTac & \textbf{79} & 59 & 59 & 190 \\
UnifySL & 677 & 713 &  \textbf{722} & 2,865 \\ 
angles & \textbf{10} & 7 & 6 & 199 \\
buchberger & \textbf{34} & 33 & \textbf{34} & 299 \\
chinese & 97 & \textbf{127} & 126 & 462 \\
\tabincell{l}{coq-library-\\undecidability} & 196 & \textbf{233} & 224 & 3,181 \\
\tabincell{l}{coq-\\procrastination} & 0 & 0 & 0 & 1 \\
coqoban & 0 & 0 & 0 & 7 \\
coqrel & 20 & 19 & \textbf{23} & 94 \\
coquelicot & 281 & 281 & \textbf{283} & 3,498 \\
dblib & 73 & 86 & \textbf{87} & 371 \\
demos & 17 & 15 & \textbf{20} & 192 \\
dep-map & \textbf{16} & 9 & 10 & 142 \\
disel & 3 & \textbf{4} & \textbf{4} & 47 \\
fermat4 & 5 & 10 & \textbf{11} & 45 \\
\tabincell{l}{fundamental-\\arithmetics} & \textbf{127} & 122 & \textbf{127} & 420 \\
goedel & \textbf{999} & 993 & 995 & 6,640 \\
hoare-tut & 3 & 3 & 3 & 27 \\
huffman & 5 & 3 & \textbf{6} & 108 \\
\tabincell{l}{jordan-curve-\\theorem} & 5,470 & 7,352 & \textbf{7,531} & 28,672 \\
tree-automata & 3,778 & 3,761 & \textbf{3,809} & 15,201 \\
verdi & \textbf{239} & 232 & 235 & 1,917 \\
verdi-raft & 1,714 & 1,747 & \textbf{1,802} & 11,063 \\
weak-up-to & 0 & \textbf{2} & 1 & 52 \\
zchinese & 40 & 47 & \textbf{52} & 247 \\
zfc & 67 & \textbf{83} & 72 & 461 \\
zorns-lemma & 337 & 337 & \textbf{360} & 2,093 \\
\midrule
Total & 14,287 & 16,278 & \textbf{16,602} & 78,494 \\
\bottomrule
\end{tabular}
\end{small}
\end{center}
\vskip -0.2in
\end{table}

\textbf{GIN vs TreeLSTM} To further evaluate the effectiveness of our encoding method, the accuracy for premise selection is also reported. We compare to the previous embedding approach using TreeLSTM as encoder $\Phi$. Of all the 3,784 instances in the PremiseGym testing set, TreeLSTM correctly selects 1,399 premises (36.98\%) for the given theorems. Our encoder successfully predicts \textbf{1,704} premises (\textbf{45.04\%}), which obtains more than \textbf{21.8\%} relative improvement on premise selection. 

\section{Conclusion}
In this work, we propose a framework using graph neural networks and graph contrastive learning for improving the representations of Coq terms for theorem proving. Our approach achieves state-of-the-art performance in the tactic prediction task. In the future, we would like to incorporate more semantic information for the representations and simplify the structure of ASTs to enhance more theorem reasoning.

\section*{Acknowledgments}
This project is supported by the Natural Sciences and Engineering Research Council (NSERC)
Discovery Grant and the Canada CIFAR AI Chair Program.

\bibliography{main}
\bibliographystyle{icml2021}

\end{document}